\documentclass[10pt,twocolumn,letterpaper]{article}

\usepackage{cvpr}
\usepackage{times}
\usepackage{epsfig}
\usepackage{graphicx}
\usepackage{amsmath}
\usepackage{amssymb}
\usepackage{diagbox}

\usepackage[breaklinks=true,bookmarks=false]{hyperref}

\cvprfinalcopy 

\def\cvprPaperID{****} 

\begin{document}

\title{Smooth Proxy-Anchor Loss for Noisy Metric Learning}

\author{Carlos Roig \ \ \ David Varas \ \ \ Issey Masuda \ \ \ Juan Carlos Riveiro \ \ \ Elisenda Bou-Balust \\
Vilynx \\
{\tt\small \{carlos,david.varas,issey,eli\}@vilynx.com}}

\maketitle

\begin{abstract}
Many industrial applications use Metric Learning as a way to circumvent scalability issues when designing systems with a high number of classes. Because of this, this field of research is attracting a lot of interest from the academic and non-academic communities. Such industrial applications require large-scale datasets, which are usually generated with web data and, as a result, often contain a high number of noisy labels. While Metric Learning systems are sensitive to noisy labels, this is usually not tackled in the literature, that relies on manually annotated datasets.

In this work, we propose a Metric Learning method that is able to overcome the presence of noisy labels using our novel Smooth Proxy-Anchor Loss. We also present an architecture that uses the aforementioned loss with a two-phase learning procedure. First, we train a confidence module that computes sample class confidences. Second, these confidences are used to weight the influence of each sample for the training of the embeddings. This results in a system that is able to provide robust sample embeddings.

We compare the performance of the described method with current state-of-the-art Metric Learning losses (proxy-based and pair-based), when trained with a dataset containing noisy labels. The results showcase an improvement of 2.63 and 3.29 in Recall@1 with respect to MultiSimilarity and Proxy-Anchor Loss respectively, proving that our method outperforms the state-of-the-art of Metric Learning in noisy labeling conditions.

\end{abstract}

\section{Introduction}
Recent deep learning applications use a semantic distance metric, which enables applications such as face verification \cite{facenet, arcface}, person re-identification \cite{chen2017triplet, xiao2016joint}, few-shot learning \cite{qiao2019transductive, sung2017learning}, content-based image retrieval \cite{proxy-nca, liftedStructure, npair_loss} or representation learning \cite{zagoruyko2015learning, proxy-nca}. These type of applications rely on vectorial spaces (embedding spaces), which are generated with the objective of gathering together the samples of the same classes while distancing themselves from the ones of other classes. 

The field of research focused on these topics is called metric learning, and it can be separated in two big categories according with the strategy used when computing the loss. Namely, this groups are \emph{proxy}-based and \emph{pair}-based losses.

\begin{figure}[t]
\begin{center}
    \includegraphics[width=0.75\linewidth]{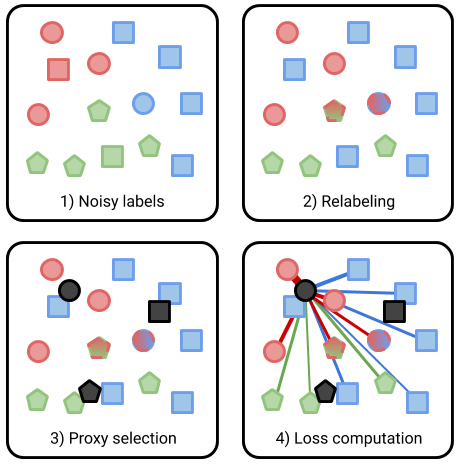}
\end{center}
   \caption{Smooth Proxy-Anchor Method. 1) Noisy labels. 2) Samples are relabeled according to class confidences. 3) Then, the proxies are selected for the relabeled samples. 4) And, finally, the loss is smoothly weighted by the class confidences.}
\label{fig:intro_image}
\end{figure}
The most important difference between proxy and pair-based losses is that proxy-based losses use embeddings placeholders as class representatives when computing similarities between samples, while pair-based losses compute similarity metrics from the vectorial representations of specific samples. Pair-based methods are more sensitive to noisy samples than proxy-based ones, since the proxy is able to ameliorate the effect of such samples. Therefore, proxy-based methods are more robust to noise.

There is a lack of literature analyzing the performance of proxy-based methods in noisy datasets. Most of research work usually tackles proxy-based losses in manually curated datasets  such as CUB-200-2011 \cite{cub-200}, Cars-196 \cite{cars-196} or Stanford Online Products (SOP) \cite{sop}. While this is sometimes possible in research, the generation of manually curated datasets in the industry is usually to expensive or unfeasible. In fact, many datasets are generated by crawling data from the Internet or other sources, resulting in a high number of noisy samples \cite{roig2020unsupervised}.

In this work we analyze the effect of noise samples in proxy-based systems, and propose an approach that is not only able to cope with noise, but in fact can use these samples to improve the performance of a metric learning system. This is achieved by modifying the state-of-the-art Proxy-Anchor loss \cite{proxy-anchor} method. Our approach extends the loss function by using class probabilities, which are used for proxy selection and sample contribution weighting. This is implemented by adding a confidence module to the embedding network, which computes the individual class probabilities by using a multi-class classification objective. In order to generalize to unseen classes, this confidence module is removed at inference time. Therefore, it is only used in training stages of the system, resulting in a network with neither computational nor size overhead. The resulting modified loss is called \emph{Smooth Proxy-Anchor Loss}. This results in a method that has the benefits of proxy-based losses while being able to benefit from noisy samples. At the same time, if a sample can not be matched to any class, it is used to improve the embedding space by separating it from other classes. 

The main contribution of this work consists on a novel proxy-based loss for metric learning that makes use of noisy samples to improve the performance of the system. This is done by modifying a state-of-the-art proxy-based loss and smoothly weighting the individual contribution of each sample to the loss, resulting in the Smooth Proxy-Anchor Loss. Moreover, we introduce a training architecture that uses two branches to reduce the impact of noisy samples, trained with the aforementioned loss. 

The rest of the paper is structured as follows: First, in Section \ref{sec:related}, we review previous works in the area of metric learning. Second, in Section \ref{sec:our_method}, we introduce our Smooth Proxy-Anchor loss and we propose a system that benefits from noisy samples. Then in Section \ref{sec:experiments}, we review the experiments that have been performed to assess our system. Finally, conclusions are presented in Section \ref{sec:conclusion}.

\section{Related Work}
\label{sec:related}
In this section we review different types of losses for metric learning, starting with those pair-based. Then, we review the literature about proxy-based losses. We end this section reviewing other works that are related with noisy metric learning.

\subsection{Pair-based Loss}
Contrastive \cite{contrastive_loss} and Triplet loss \cite{facenet} have been key in the development of the field of metric learning. The objective of both losses is to group together samples of the same class, while pushing away samples of other classes. For the Contrastive loss, only pairs of samples are used. Instead, the Triplet loss uses three samples, two from the same class (anchor and positive) and one from another class (negative).
N-pair loss \cite{npair_loss} and Lifted Structure loss \cite{liftedStructure} generalized the Triplet loss to a pair of samples from the same class against many negative ones, by pulling together the positive pair and pushing away all the negatives. These methods do not use the complete information in a batch, because they select a predefined number of samples for each pair. To overcome this limitation, Ranked List loss \cite{ranked_list_loss} proposes a method that takes into account all the samples in a batch by trying to separate the positive and negative samples. MultiSimilarity loss \cite{Multisimilarity-loss} goes one step further by weighting the influence of each pair in the batch, trying to focus on more useful pairs and resulting on a performance and speed gain.
The complexity of pair-based methods grow with the number of tuples of data considered in a batch. Different works \cite{facenet, sampling_matters} analyze the impact of having a big number of tuples, leading to a performance reduction. To solve the increment in tuples, different mining approaches have been proposed in the literature \cite{facenet, sampling_matters, smart_mining}, reducing the complexity of the problem and improving the performance of the generated embeddings. 
\subsection{Proxy-based Loss}
Proxy-based methods emerged with Proxy-NCA loss \cite{proxy-nca} in order to cope with the complexity problem. By generating a proxy embedding for each of the classes in the dataset, these methods compute the distance between the samples in the batch and the proxies. This reduces the complexity (lowers the number of possible combinations) and is more robust to noisy samples. Afterwards, SoftTriple was introduced as a modification of SoftMax loss for classification, which works on a similar way than Proxy-NCA but assigns multiple proxies for each class, resulting in a better intra-class representation. More recently, Proxy-Anchor loss has been presented as a proxy-based loss that takes advantage from pair-based methods. To do so, the loss takes into account all the samples of the batch and weights them based on their distance to the proxies.

\subsection{Noisy Metric Learning}
Proxy-based and pair-based losses are usually analyzed in manually curated datasets. In fact, there has been very few research on how noise affects metric learning losses. In \cite{robust_distance_ml_noise} an Expectation-Maximization algorithm based on Neighbourhood Component Analysis (NCA) is proposed to overcome the label noise present on datasets. Similarly, \cite{huang2012robust} formulates the metric learning task as a combinatorial optimization problem based on smooth optimization \cite{smooth_optimization}. The approaches proposed are oriented towards very small datasets (single class, 600 images), and are not suitable for larger datasets.

\section{Our Method}
\label{sec:our_method}
In this section, we first formulate the problem of noisy metric learning. Second, we review the Proxy-Anchor loss \cite{proxy-anchor}. Then, we introduce our Smooth Proxy-Anchor loss, which is based on the previous one. This loss benefits from the noisy labels in the dataset, learning a more robust embedding space. We finally propose an implementation of an architecture that is trained using the Smooth Proxy-Anchor loss. Note that this loss is not specific to any particular architecture.

\subsection{Problem Formulation}
Let us define $\mathcal{Z}$ as the data space from which we sample a set of data points $Z = [z_1, z_2, ... , z_N]$ with their associated noisy labels ${y_i} \in {1, ..., C}$, where \textit{C} represents the number of classes. These labels form the label set ${Y} = [{y_1}, {y_2}, ... , {y_N}]$. Note that, due to the nature of how the samples are obtained, even though each sample $z_i$ has its corresponding label ${y_i}$, this label may not be correct.

Let $f : \mathcal{Z}\to \mathcal{V}$ be a mapping from the data space to a given latent space, where each vector $v_i$ has semantic characteristics of its corresponding data point $z_i$. The objective of metric learning is to learn a distance metric
in this space space so that it can reflect the actual semantic
distance. The distance metric is defined as:
\begin{equation}
    D(z_i, z_j) = m(f(z_i; \theta), f(z_j; \theta))
\end{equation}
where $m$ is a positive symmetric function and $\theta$ is a vector that parameterizes the mapping function $f$.

Deep learning methods usually extract features using a deep neural network. A standard procedure is to first project
the features into an embedding space (or metric space) $\mathcal{X}$
with a mapping $g : \mathcal{V} \to \mathcal{X}$, where the distance metric is known. Since the projection can
be incorporated into the deep network, we can directly learn
a mapping $h : \mathcal{Z} \to \mathcal{X}$ from the data space to the
embedding space, so that the whole model can be trained
end-to-end without explicit feature extraction. In this case,
the distance metric is defined as:
\begin{equation}
    D(z_i, z_j) = d(x_i, x_j) = d(h(z_i; \theta_h), h(z_j; \theta_h))
\end{equation}
where $d$ indicates a given distance , $x_i = h(z_i)$ and $x_j = h(z_j)$ are the learned embeddings and $\theta_h$ are the parameters of the network. These parameters are learned by minimizing a specific loss function $\mathcal{L}\left( X \right)$. Equivalently, these distances can be learned using similarity style supervision. Then, the similarity between a pair of samples is defined as:
\begin{equation}
    s(x_i, x_j) = 1- D(x_i, x_j)
\end{equation}

\subsection{Review of Proxy-Anchor loss}
Proxy-Anchor loss is introduced in \cite{proxy-anchor} as a method to overcome the limitations of the Proxy-NCA loss \cite{proxy-nca} without increasing the training complexity. The key idea of this loss is to use each proxy as an anchor and associate it with all the batch samples. Proxy-Anchor loss is defined as:
\begin{equation}
\begin{aligned}
\mathcal{L}\left( X \right) =
& \frac{1}{ |P^{+}|}\sum_{p\in P^{+}} \textup{log}\left ( 1 + \sum_{x\in X_{p}^{+} }e^{-\alpha (s(x,p)-\delta )}\right ) \\
& + \frac{1}{|P|}\sum_{p\in P} \textup{log} \left ( 1 + \sum_{x\in X_{p}^{-} }e^{\alpha (s(x,p)+\delta  )}\right)
\end{aligned}
\label{eq:proxy_anchor_loss}
\end{equation}

where $x$ is the embedding corresponding to an image, $s(\cdot, \cdot)$ is the cosine similarity between two embedding vectors. $P$ is the set of all proxies and $P^{+}$ the positive set of proxies for a given sample in the batch. $X_{p}^{+}$ is the subset of positive samples corresponding to the proxy $p$ from the batch $X$, while $X_{p}^{-}$ is the opposite, being  $X_{p}^{-} = X - X_{p}^{+}$. Moreover, $\delta$ and $\alpha$ are hyperparameters denoting the margin and scaling factor.

The loss presented in Equation \ref{eq:proxy_anchor_loss} aims to pull $p$ and its
most dissimilar positive example (hardest positive example) together, and to push $p$ and its most similar negative example (hardest negative example) apart. In practice, Proxy-Anchor pulls/pushes all embedding vectors in the batch with different strength according to their relative hardness. 

Although proxy based methods are more robust to noisy labels than other techniques, these errors still affect their performance. This is specially critical when embeddings from different classes are pulled together, as their proxies are moved away from their optimal position. This results in a negative effect on all the other samples of their class.

\subsection{Smooth Proxy-Anchor Loss}

\begin{figure*}
\begin{center}
\includegraphics[width=1\linewidth]{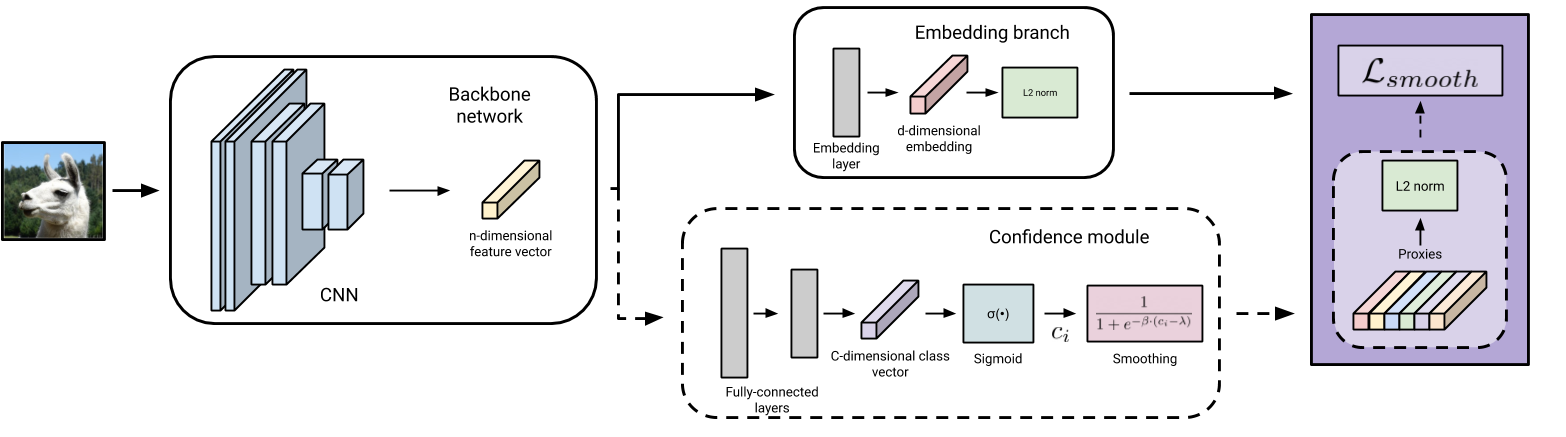}
\end{center}
   \caption{Our proposed architecture composed of three main blocks. The backbone networks extracts feature vectors that are fed to the two following blocks. The confidence module generates class weights, while the embedding branch generates an embedding of the input sample. Finally, the loss is computed comparing the embedding with the proxies, with a weighting based on the class confidences.}
\label{fig:architecture}
\end{figure*}

\label{sec:loss}
In order to overcome the problems associated with noisy samples, we propose a modified Proxy-Anchor loss that takes into account the probability of each sample to belong to a given class. Using this method, we benefit from the noisy nature of the labels to probabilistically pull/push samples and proxies during the training process. As in Proxy-NCA \cite{proxy-nca} and Proxy-Anchor \cite{proxy-anchor}, our loss assigns a single proxy to each class following the standard proxy assignment setting.
We define our Smooth Proxy-Anchor Loss as:
\begin{equation}
\begin{aligned}
& \mathcal{L}_{smooth}\left( X \right) = \\
& \frac{1}{ |P^{+}|}\sum_{p\in P^{+}} \textup{log}\left ( 1 + \sum_{x\in X_{p}^{+} }w_{x,p}\cdot e^{-\alpha (s(x,p)-\delta )}\right ) \\
& + \frac{1}{|P|}\sum_{p\in P} \textup{log} \left ( 1 + \sum_{x\in X_{p}^{-} }(1-w_{x,p})\cdot e^{\alpha (s(x,p)+\delta  )}\right)
\end{aligned}
\label{eq:smooth_proxy_anchor}
\end{equation}
where ${w}_{x,p} \in [0,1]$ represents a weighting smooth function associated with the confidence of $x$ to belong to the same class as the proxy $p$. 

We also propose a new method to select positive/negative proxies for each sample in the batch. As in \cite{proxy-anchor}, for each proxy
$p$, a batch of embedding vectors $X$ is divided into two sets
$X^{+}_p$ and $X^{-}_p$ such that $X = X^{+}_p + X^{-}_p$. However, we propose to select the set of positive samples $X^{+}_p$ for each proxy $p$ as those samples with a confidence of belonging to the $i_{th}$ class above a given threshold $\lambda$. The remaining samples in the batch are included in the negative set $X^{-}_p$. This threshold allows us both to assign each sample to more than one proxy and to ensure that this sample does not belong to the  positive and negative proxy sets at the same time.

Two major differences with previous works arise from this selection of positive/negative samples for each proxy. First, the original labels $y_i$ are not directly used for learning our distance. Instead, a confidence value associated with each class is used. Second, each sample may belong to more than one positive set of proxies. In other words, samples of the batch can be pulled towards more than one proxy at the same time. Moreover, they are pulled with different weights depending on their class confidence. These two characteristics make our method robust against noisy labels and allow us to benefit from the nature of those images that are close to more than one class.

As in \cite{proxy-anchor}, our loss pulls and
pushes all embedding vectors in the batch with different strength that is associated with their similarity. However, our method allows us to balance the influence of each sample according to its class confidences. This can be observed in the gradient of our loss with respect to $s(x,p)$:
\begin{equation}
    \frac{\partial \mathcal{L}_{smooth}}{\partial s(x,p)}= 
    \begin{cases}
    \frac{w_{x,p}}{|P^{+}|} \frac{-\alpha h^{+}_p(x)}{1+\hspace{-0.23cm}\sum\limits_{x' \in X^{+}_p}h^{+}_p(x')},& \forall x \in X^{+}_p\\
    \frac{(1-w_{x,p})}{|P|} \frac{\alpha h^{-}_p(x)}{1+\hspace{-0.23cm}\sum\limits_{x' \in X^{-}_p}h^{-}_p(x')},& \forall x \in X^{-}_p
    \end{cases}
    \label{eq:gradient_loss}
\end{equation}
where $h^{+}_{p}(x) = e^{- \alpha (s(x,p) - \delta)}$ and $h^{-}_{p}(x) = e^{ \alpha (s(x,p) + \delta)}$ are positive and negative hardness metrics for embedding vector $x$ given proxy $p$, respectively. The scaling parameters $\alpha$ and margin $\delta$ control the relative hardness of data points. As a consequence, they determine how strongly pull or push their embedding vectors.

As it can be observed in Equation \ref{eq:gradient_loss}, the gradient becomes larger when $x$ is harder than the others samples. Moreover, this gradient is scaled with a weight associated with the confidence of each proxy class:
\begin{equation}
\label{eq:smoothing}
    {w}_{x,p} = \frac{1}{1+e^{-\beta \cdot(c_{x,p}-\lambda)}}
\end{equation}
where $c_{x,p}$ is the confidence value for sample $x$ of belonging to the same class than $p$. The parameters $\beta$ and $\lambda$ control the sharpness and the position of the smoothing function. Note that $\lambda$ also defines $X^{+}_{p}$ and $X^{-}_{p}$ for each proxy $p$, as discussed before.

Using these weights, the contribution to the loss of positive samples is reduced if they have low class confidence whereas the gradient of negative samples is reduced as their confidence of belonging to the proxy class increases.

\subsection{Architecture for Smooth Proxy-Anchor Loss}
\label{sec:architecture}

In this section, we propose a system architecture in which the loss presented in Section \ref{sec:loss} is used for generating robust embeddings from images with noisy labels. As it can be observed in Figure \ref{fig:architecture}, this architecture is composed by three main blocks: a backbone used as a feature extractor, an embedding branch and a confidence extraction module. The first two blocks define the basic structure of embedding architectures, while the confidence module is added to make these embeddings robust against noise. The parts of the architecture that are only used for training the system are marked with dashed lines in the figure. 

The training of the entire system is divided in two phases. In the first phase, the confidence module is trained for multi-class classification using the noisy labels. During the second phase of the training, the backbone generates a feature vector that is used both to extract an embedding of the image and to perform multi-class classification with the frozen confidence module. Equation \ref{eq:smooth_proxy_anchor} shows how we use this confidence to compute the loss. At inference time, the features generated by the backbone are only used to compute the embedding of the image. 

The embedding branch is formed by a single embedding layer in order to keep the network as simple as possible. This layer is trained using our Smooth Proxy-Anchor loss with the class confidences extracted from the multi-class classification layer.

The confidence module is composed of two fully-connected layers. The first layer is used to reduce the dimensionality of the features extracted from the backbone network, while the classification layer generates class confidences. As this module is trained for multi-class classification, its output is a $C$-dimensional vector, where $C$ is the number of different classes in the training set.

The confidence module is not used for inference because the confidences can only be computed over the training classes. In other words, the confidence module improves the results during training making the system robust against noise in order to obtain a feature space with better generalization over unseen classes. During inference, our system results in a backbone architecture and single embedding layer that generates feature vectors.

\subsection{Training in two phases using noisy labels}
\label{sec:two_trainings}
As previously exposed, the training of our system is a two-phase training procedure. We discuss the motivation of such approach in this section. 

In the first training phase, the confidence module is trained with a multi-class classification objective. In such scenario, the network tries to identify the common patterns in the input data that belong to the same class. Mislabeled samples in the dataset behave like outliers for the model and, thus, the model is not able to classify them correctly.


During the second training phase, we set the confidence module in inference mode. This means that it does not train together with the embedding network. Instead, it only provides class confidences. Here is where we make use of the fact that the confidence module will not classify mislabeled samples as the ground truth class. Leveraging such information is what allows our embedding network trained with the proposed Smooth Proxy-Anchor Loss to overcome its predecessors. Furthermore, the output of confidence module is not a binary \{0,1\} prediction per class, but a continuous value. This output, then, contains compressed information about the distance between a sample and each class. The Smooth Proxy-Anchor Loss uses this information to weight sample contributions in a  fine-grained manner.

The key idea behind this training strategy is to replace the original noisy labels with class confidences for the training of our embeddings. Even though this mislabeled sample counts as an error for the classification task during the first training, the incorrect labeling proves to be beneficial when we consider its class confidence during the second training phase, the embedding training. In particular, a noisy label associated with an image is translated into a low class confidence. This results either in a low contribution of the sample if it is classified as positive for its class (confidence higher than $\lambda$) or a large contribution if it is classified as negative (confidence lower than $\lambda$). Note that both cases are beneficial for obtaining better embeddings.

This strategy relies on the assumption that no  overfitting is present during training of the confidence network. Otherwise, this module would memorize these mislabeled samples from the training set.

\section{Experiments}
\label{sec:experiments}
We introduce in this section the setup for our experiments and the different experiments done to show the benefits of using the Smooth Proxy-Anchor Loss.
We start with the review of the dataset. Then, we move on to the specific details of the implementation. Afterwards, we present the experiments carried out to evaluate the performance of the multi-class classification block. Finally, we show how noise affects metric learning losses and review how our proposed loss performs against these losses.

\subsection{Dataset}
\label{sec:dataset}
Due to the lack of standardization in noisy datasets for deep metric learning, we propose to use a subset of the WebVision dataset \cite{webvision} in order to conduct our experiments. The WebVision dataset has more than 16 million images from 5000 classes, crawled from different image search engines. The dataset is designed to facilitate the research on learning visual representations from noisy web data. 

Following the commonly used data split protocol for deep metric learning \cite{liftedStructure}, we select the first 100 classes of the WebVision dataset for training. We set the maximum number of samples per class at 300 in order to prevent class imbalance, resulting in a total of 29707 training images.

For testing purposes, we generate two splits. We use these splits in the assessment of the two main tasks performed in this work. For the first phase, multi-class classification, we select a maximum of 100 samples of each class present in the training set. In order to ensure clean samples and avoid repeating training images for the validation set, we select these samples from the WebVision validation split, resulting in a selection of 9727 images. For the second phase, that is composed of the metric learning experiments, we also use the validation split of the WebVision dataset, but in this case, the samples that are selected correspond to 100 unseen classes (101 to 200 of the dataset), which amounts to 7382 images. This is the standard procedure to assess both accuracy and generalization in deep metric learning tasks for computer vision \cite{proxy-nca, Multisimilarity-loss, proxy-anchor}.

Note that the size of our subset and the number of classes are comparable or even larger than standard datasets of deep metric learning \cite{cars-196, cub-200}, which do not contain noisy labels.

\subsection{Implementation Details}
In this section we present the implementation specifics of our system. We start reviewing the architecture proposed, which is composed of three blocks, namely the backbone, confidence module, and embedding head. Then we detail the specifics of how the blocks are trained and optimized, and we conclude this section with a discusion of the hyperparameters of the Smooth Proxy-Anchor Loss. 

\subsubsection{Architecture}
Our proposed system is composed of three main blocks: the backbone feature extractor, embedding head and confidence module. We individually review the implementation details of each of the three block.

\textbf{Backbone network:} The architecture selected as feature extractor is a ResNet50 \cite{resnet}, which represents a good trade-off between complexity and accuracy in classification tasks. However, any other architecture could be used as backbone instead of it. This network is used without the last layer, pre-trained on the ImageNet dataset \cite{imagenet_cvpr09} and frozen for all the experiments performed in our work. Note that any other dataset could be used for pre-training this network. We choose this configuration because it is commonly used in a wide variety of computer vision tasks. 

\textbf{Embedding head:} We set the feature embedding size to 64 for all the experiments, and we normalize features using L2-normalization. The weights of the embedding layer are initialized as in \cite{he_init}, while the bias are set to zero.

\textbf{Confidence module:} In order to perform multi-class classification, we use two fully-connected layers. The first layer reduces the output feature of the backbone from 2048 to 512, and it is followed by a ReLU activation function. The classification layer has an output size 100 (number of training classes), with sigmoid activation function. This layer is initialized using Xavier initialization method \cite{glorot_uniform}.

\subsubsection{Training}
As we have discussed in Section \ref{sec:two_trainings}, we train the complete system in two separate phases. In the first phase, we train the confidence module that is used for the multi-class classification task, while in the second phase, we learn the embeddings for the noisy metric learning task using the class confidences provided by the frozen confidence module.

We use data augmentation to prevent overfitting. The same procedure is used for both tasks. We perform random cropping and horizontal flipping for training and only center-cropping during testing. The inputs are
resized to 256 by 256 pixels, and then randomly cropped resulting in images of size 224 by 224. We randomly select the input images for each batch during training.

During the first training phase, we tackle the training of the confidence module as a multi-class classification task. Thus, the classification layer is trained using a standard binary cross entropy loss with sigmoid activation function.
During the second phase, we train the embedding branch using our Smooth Proxy-Anchor Loss.
Both stages are trained with the Adam optimizer with weight decay \cite{weight_decay} and a learning rate of $10^{-4}$ for all the dense networks (embedding, dimensionality reduction and classification) and $10^{-2}$ for proxies. 

\subsubsection{Hyperparameters}
The hyperparameters of the Proxy-Anchor loss, $\alpha$ and $\delta$, are set to 32 and $10^{-1}$ respectively for all the experiments. These are the values used by the authors of this loss in \cite{proxy-anchor}. On the other hand, the hyperparameters associated to our weighting function presented in Equation \ref{eq:smoothing}, $\beta$ and $\lambda$, are set to  100 and $10^{-1}$ respectively. We detail the procedure to select these values in Section \ref{sec:smooth_proxy_anchor_loss}.

\subsection{Multi-Class Classification}

\begin{figure*}
\begin{center}
\includegraphics[width=0.9\linewidth]{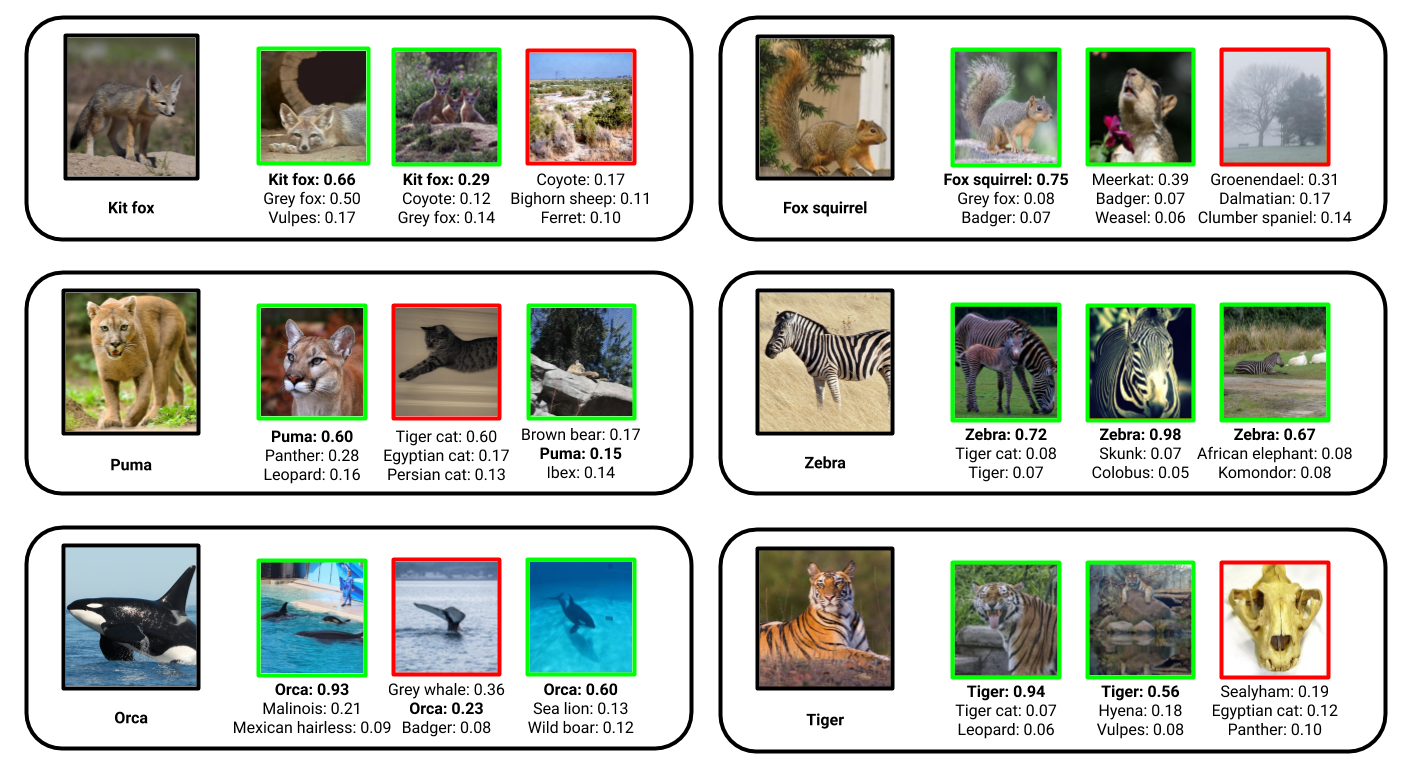}
\end{center}
   \caption{Results of multi-class classification when trained with noisy samples. Each set of images is composed with a class example image (black border), and three images from the dataset. The dataset images are represented with a \textcolor{green}{green} border if the label is correct or \textcolor{red}{red} border if the sample is incorrectly annotated. We also show the top-3 accuracy for each sample, remarking with \textbf{bold} their class label.}
\label{fig:multiclass}
\end{figure*}

We analyze the presence of noise and how it affects the training of a network, when it is trained on a multi-class classification objective. The experiment is carried out training the confidence module proposed in Section \ref{sec:architecture}, achieving an accuracy of 72.56\% on the classification test set.

In Figure \ref{fig:multiclass} we show how this classification branch of the network performs when trained with noisy samples. In this figure, a green border around an image marks it to be correctly labeled, while a red border means otherwise. Moreover, we present the top-3 confidences returned by the network and their corresponding labels.

The results show that the network is able to identify with high confidence correct samples, while giving low confidence values to image that do not have a proper depiction of the class (e.g. third sample of class \textit{Kit fox}). We also see that samples labeled with another class of the dataset are correctly classified by our confidence module (e.g. second sample of class \textit{Puma}).

\subsection{Noisy Metric Learning}
In this section, we evaluate the impact of noisy samples in the dataset for the task of deep metric learning. First, we analyze the behaviour of two common proxy-based losses in this scenario. Then, we assess our proposed loss and we show how noisy labels are used to improve the performance using noisy labels. Finally, we show a comparison between our loss and two state-of-the-art pair-based and proxy based methods.

\subsubsection{Noise Relevance}
\label{subsec:noise_relevance}
In order to analyze the impact of noisy samples on the performance of proxy-based losses, we propose an experiment in which we train an embedding branch using Proxy-NCA and Proxy-Anchor losses with the noisy training dataset presented in Section \ref{sec:dataset}. 

The main idea of this experiment is to perform two trainings for each loss. We conduct the first training using all the samples of the train dataset. In the second training, we do not consider those samples with low class confidence for their label. These samples are considered to be noisy if the confidence assigned to the class specified by their label is smaller than a threshold $\lambda_c$. This class confidence is estimated using our confidence module and the $\lambda_c$ is set to $10^{-1}$ for this experiment.

\begin{table}
\begin{center}
\begin{tabular}{|l|c|}
\hline
Method & Recall@1 \\
\hline\hline
Proxy-NCA & 65.89 \\
\hline
Proxy-NCA w/o noise & 66.2 \\
\hline
Proxy-Anchor & 67.95 \\
\hline
Proxy-Anchor w/o noise & \textbf{68.14} \\
\hline
\end{tabular}
\end{center}
\caption{Comparison of proxy-based losses with and without noisy samples.}
\label{table:noise_relevance}
\end{table}

Table \ref{table:noise_relevance} shows that, although Proxy-NCA and Proxy-Anchor are proxy-based, the noise present in the dataset affects the performance of both losses. This can be observed in the improvement of the results corresponding to both losses when we ignore the aforementioned noisy samples (Proxy-NCA w/o noise and Proxy-Anchor w/o noise). 

As the noisy samples of the dataset are not labeled as such, we cannot perform further analysis on the behaviour of these losses varying the ratio of noisy labels without manual annotation. We set a conservative value of $\lambda_c$ to ensure that the minimum number of positive images are removed from the train dataset. Using this threshold in our experiment, we show that both losses decrease their performance in the presence of noisy labels.

\subsubsection{Smooth Proxy-Anchor Loss}
\label{sec:smooth_proxy_anchor_loss}

\begin{table}
\begin{center}
\begin{tabular}{|l|c|c|c|c|}
\hline
\backslashbox{$\lambda$}{$\beta$} & 50 & 100 & 150 & 200 \\
\hline\hline
0.075 & 69.79 & 69.94 & 69.86 & 69.8 \\
\hline
0.1 & 71.15 & \textbf{71.24} & 71.15 & 71.09 \\
\hline
0.125 & 70.74 & 70.48 & 70.62 & 70.66 \\
\hline
0.15 & 70.05 & 70.12 & 70.06 & 70.04 \\
\hline
\end{tabular}
\end{center}
\caption{Comparison of Smooth Proxy-Anchor Loss when modifying the hyperparameters $\lambda$ and $\beta$.}
\label{tab:hyperparams}
\end{table}

In this experiment, we assess the performance of our Smooth Proxy-Anchor Loss with a dataset affected by noise. Moreover, we also investigate how the hyperparameters $\lambda$ and $\beta$ of the smoothing function (Equation \ref{eq:smoothing}) affect the quality of the embeddings generated by our system. 

We propose to analyze the impact of these hyperparameters for several pairs of values. In Table \ref{tab:hyperparams}, we show how the different values of $\lambda$ and $\beta$ influence the performance of our loss. As it is shown in Table \ref{tab:loss_comparison}, our loss outperforms \cite{Multisimilarity-loss, proxy-anchor} with any pair of values presented in the previous table. As samples with a class confidence larger than $\lambda$ are considered as positive for such class, when this parameter is too low, it can result in samples considered to belong to too many classes. On the other hand, if $\lambda$ is too high, the most informative samples (hard positives) may not have any label assigned and could be pushed away from all the proxies. 
The value of $\beta$ smooths the contribution of each sample to the batch relying on its confidence, being a soft pull/push for small values while becoming a hard one for bigger values of $\beta$.

\subsubsection{Comparison with other methods}

\begin{table}
\begin{center}
\resizebox{\linewidth}{!}{
\begin{tabular}{|l|c|c|c|c|c|}
\hline
Recall@K & 1 & 2 & 4 & 8 & 16 \\
\hline\hline
Proxy-NCA \cite{proxy-nca} & 65.89 & 75.70 & 82.36 & 87.51 & 91.56 \\
\hline
Proxy-Anchor \cite{proxy-anchor} & 67.95 & 77.47 & 84.50 & 89.33 & 93.09 \\
\hline
MultiSimilarity \cite{Multisimilarity-loss} & 68.61 & 70.08 & 85.04 & 89.95 & 93.42 \\
\hline
\textbf{Ours} & \textbf{71.24} & \textbf{79.83} & \textbf{86.10} & \textbf{90.30} & \textbf{93.66} \\
\hline
\end{tabular}}
\end{center}
\caption{Comparison of Recall@K for different methods against our proposed loss.}
\label{tab:loss_comparison}
\end{table}
In order to demonstrate that our loss is robust against noisy labels in the dataset, we compare our results with two state-of-the-art losses: Multisimilarity \cite{Multisimilarity-loss} loss and Proxy-Anchor \cite{proxy-anchor} loss. These losses are pair-based \cite{Multisimilarity-loss} and proxy-based \cite{proxy-anchor} respectively. The performance is measured using the standard Recall@$K$ metric, where $K$ $\in$ \{1, 2, 4, 8, 16\}. Table \ref{tab:loss_comparison} shows that our proposed loss outperforms both methods under the presence of noisy samples for any value of $K$. Moreover, as discussed in Section \ref{sec:smooth_proxy_anchor_loss}, our loss improves the results of \cite{Multisimilarity-loss} and \cite{proxy-anchor} for any pair $(\lambda, \beta)$ in Table \ref{tab:hyperparams}, with a Recall@1 increase ranging from $1,18$ to $3,29$ points.

As shown in Section \ref{subsec:noise_relevance}, both Proxy-NCA and Proxy-Anchor losses show better performance when samples with noisy labels are removed from the dataset. Although both are proxy-based losses, their Recall@1 is slightly lower than the result obtained by the Multisimilarity loss in the task of noisy metric learning. Compared to these methods, our loss demonstrates to be more robust when noisy labels are present in the dataset.

\section{Conclusion}
\label{sec:conclusion}
In this work, we have proposed a system to tackle the problem of noisy metric learning using a novel Smooth Proxy-Anchor loss. This system consists on a double branched architecture, where the first branch -confidence module- computes sample class confidences and the second one -embedding head- generates the sample embeddings. The proposed Smooth Proxy-Anchor Loss uses the class confidence scores generated by the confidence module to weight the contribution of each sample, in order to improve performance in the presence of noisy labels.

The training of this system is done in two stages: first, the confidence module is optimized with a multi-class classification objective. Then, this branch is frozen to train the embedding module using our Smooth Proxy-Anchor Loss.

Our proposed method outperforms other state-of-the-art losses when dealing with a noisy dataset: improving 2.63 and 3.29 in Recall@1 with respect to MultiSimilarity and Proxy-Anchor loss respectively. Moreover, our training strategy consisting on replacing the original noisy labels with class confidence, is able to take benefit of incorrect labeling to obtain more robust embeddings.


{\small
\bibliographystyle{ieee_fullname}
\bibliography{egbib}
}

\end{document}